\documentclass[sigconf]{acmart}

\usepackage{booktabs} 

\setcopyright{rightsretained}

\acmArticle{27}

\copyrightyear{2021}
\acmYear{2021}
\setcopyright{acmlicensed}\acmConference[ICVGIP '21]{Indian Conference on
Computer Vision, Graphics and Image Processing}{December 19--22,
2021}{Jodhpur, India}
\acmBooktitle{Indian Conference on Computer Vision, Graphics and Image
Processing (ICVGIP '21), December 19--22, 2021, Jodhpur, India}
\acmPrice{15.00}
\acmDOI{10.1145/3490035.3490284}
\acmISBN{978-1-4503-7596-2/21/12}
\editor{Chetan Arora}
\editor{Parag Chaudhuri}
\editor{Subhransu Maji}

\begin{document}

\title{Intelligent Video Editing: Incorporating Modern Talking Face Generation Algorithms in a Video Editor}

  \author{Anchit Gupta}
  \authornote{Both the authors have contributed equally to this research.}
  \affiliation{%
    \institution{IIIT, Hyderabad}
    \state{Telangana}
    \country{India}
    \postcode{500032}
  }
  \email{anchit.gupta@research.iiit.ac.in}
  
  \author{Faizan Farooq Khan}
  \authornotemark[1]
  \affiliation{%
    \institution{IIIT, Hyderabad}
    \state{Telangana}
    \country{India}
    \postcode{500032}
  }
  \email{faizan.farooq@students.iiit.ac.in}
  
  \author{Rudrabha Mukhopadhyay}
  \affiliation{%
    \institution{IIIT, Hyderabad}    
    \state{Telangana}
    \country{India}
    \postcode{500032}
  }
  \email{radrabha.m@research.iiit.ac.in}

  \author{Vinay P. Namboodiri}
  \affiliation{%
    \institution{University of Bath}
    \country{England}
    \postcode{500032}
  }
  \email{vpn22@bath.ac.uk}
  
  \author{C. V. Jawahar}
  \affiliation{%
    \institution{IIIT, Hyderabad}
    \state{Telangana}
    \country{India}
    \postcode{500032}
  }
  \email{jawahar@iiit.ac.in}  

\renewcommand{\shortauthors}{}

\begin{teaserfigure}
  \includegraphics[width=\textwidth,height=230px]{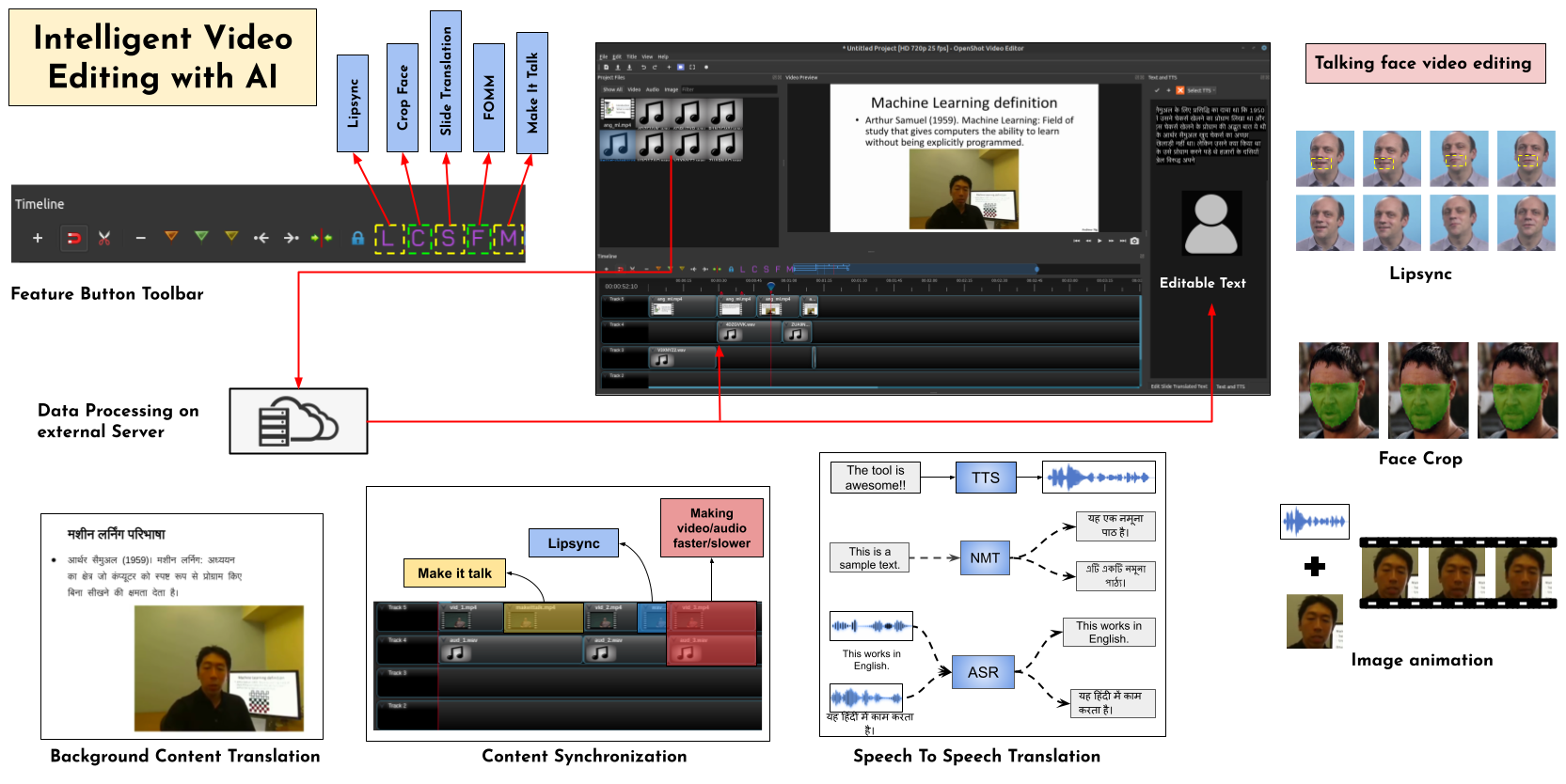}
  \caption{Our video editor aims to bring together the latest advancements in talking face generation algorithms in an interactive manner. The tool provides ample manual control over several aspects of these algorithms leading to better quality video generation. Our tool can be used extensively to translate lectures and dub movie scenes in various languages.}
  \label{fig:arch}
  \vspace{0.3cm}
\end{teaserfigure}

\begin{abstract}
This paper proposes a video editor based on OpenShot with several state-of-the-art facial video editing algorithms as added functionalities. Our editor provides an easy-to-use interface to apply modern lip-syncing algorithms interactively. Apart from lip-syncing, the editor also uses audio and facial re-enactment to generate expressive talking faces. The manual control improves the overall experience of video editing without missing out on the benefits of modern synthetic video generation algorithms. This control enables us to lip-sync complex dubbed movie scenes, interviews, television shows, and other visual content. Furthermore, our editor provides features that automatically translate lectures from spoken content, lip-sync of the professor, and  background content like slides. While doing so, we also tackle the critical aspect of synchronizing background content with the translated speech. We qualitatively evaluate the usefulness of the proposed editor by conducting human evaluations. Our evaluations show a clear improvement in the efficiency of using human editors and an improved video generation quality. We attach demo videos with the supplementary material clearly explaining the tool and also showcasing multiple results.
\end{abstract}

%


\maketitle

\section{Introduction}   
\begin{figure*}[h]
  \includegraphics[width=\textwidth, height=190px]{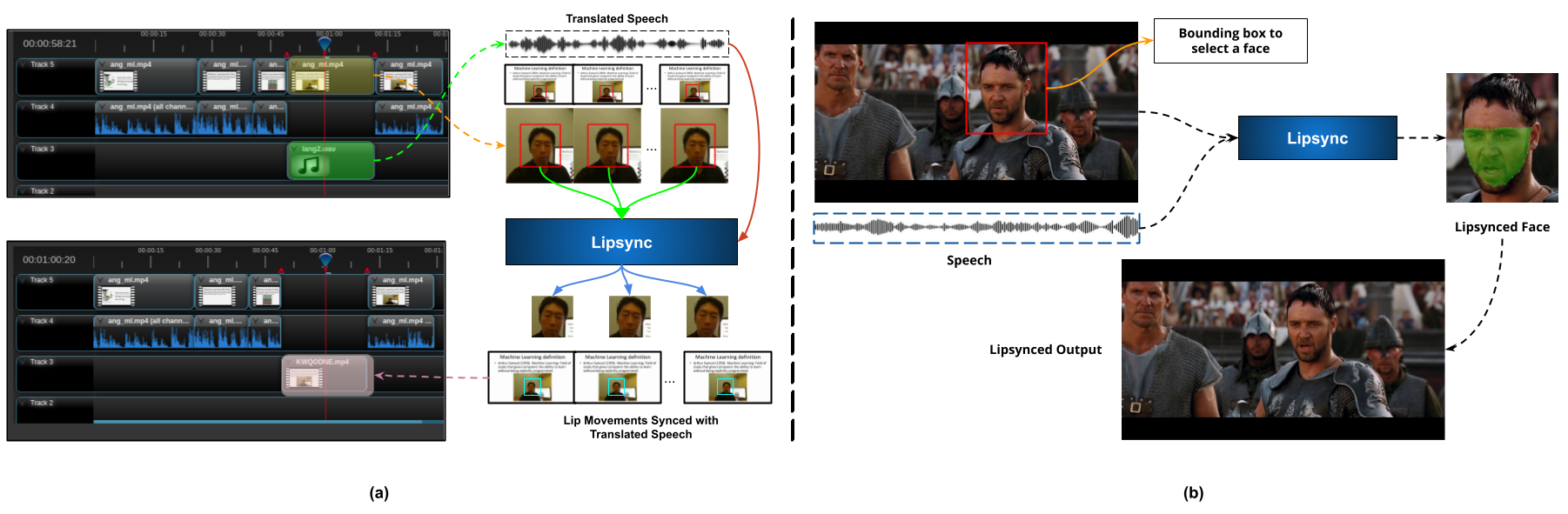}
  \caption{Demonstration of the Lip Sync feature in the tool.. Figure (a) shows the video segment selected by the user to lip synchronize. Figure (b) shows the selection of a particular face in a frame to lip synchronize and the replacement of lip synchronized face region encased by the facial key points in the original frame.
}
  \label{fig:wav2lip_fig}
  \vspace{-0.3cm}
\end{figure*}
 
In today's world, humans are more connected than ever due to social networking phenomena leading to tremendous opportunities for information sharing. Advancement in internet connectivity has led to a massive increase in the consumption of online video materials too. These include movies, talk shows, educational lectures, and vlogs. Most of this content is present in languages like English, making the content less popular among the local populace across countries. 

Dubbing is the most popular method to make the content available for consumption in different languages, but it produces unnatural visual content due to out-of-sync lip movements. Our work introduces the users to an easy-to-use framework built on an open-source video editing tool OpenShot~\cite{openshotstudios}, to reduce the manual effort while correcting out-of-sync lip movements of complex movie scenes. We add video editing features to lip-sync the speaker's lip movements according to the dubbed speech. In the presence of multiple speakers, our tool allows the human editor to select the speaker to lip-sync manually. We avoid artifacts by using facial key-point detection to replace any remnant out-of-sync lip movements. We allow the users to manually remove any undesirable frame or chunk of the video according to their discretion. Additional features for the generation of expressive talking faces from only audio and facial reenactment are present in the tool, providing users with creative ways to edit videos. The manual control in the tool helps in both the experience of editing and generating better quality results.

On the unavailability of dubbed speech, a pipeline to generate translated speech was proposed in~\citet{lipgan}. However, the automatic pipeline had severe limitations due to a lack of manual control. This motivated us to bring together the best from translation and talking face features in a tool that incorporates the automatic features of a face-to-face translation system with controllable manual editing.  To translate the original speech, we use an automatic-speech-recognizer(ASR)~\cite{SileroModels} followed by neural-machine-translation(NMT) systems~\cite{jerin1} to get the translated text. Speech is generated using the text-to-speech algorithm~\citet{glowtts} from translated text. The easy-to-use interfaces allow the user to correct transcription and translation errors. 

Out-of-sync problems arise when replacing the original speech with the automatically translated speech. This out-of-sync content often gets cumulated when translating long videos, and automatic techniques like~\citet{lipgan} prove inadequate in such scenarios. We, therefore, propose various methods of synchronizing content, discussed in \autoref{Content Sync}, helping to preserve the quality of the lecture. We also use this opportunity to include an additional feature overlooked in~\cite{lipgan} by automatically translating the background text. Similar to other features, the background translation also allows for manual intervention to improve the accuracy of the translated content. Our tool can help create translated versions of famous courses in Indian languages, making an ever-lasting impact. 

We do extensive human evaluations to prove the effectiveness of the tool. Demo videos are included in the supplementary material to clearly explain the process of editing videos using the tool and showcasing multiple results. 

\section{Current Deep Learning based tools}

Text2Vid~\cite{yao2021talkinghead} is a text-based tool for editing talking-head videos by manipulating the wording of the speech and non-verbal aspects of the performance by inserting mouth gestures like ``smile'' or changing the overall performance style. DeepFaceLab~\cite{perov2021deepfacelab} and FaceSwap~\cite{faceswap} are open-source deep learning-based video editing tools for achieving photorealistic face-swapping results. Yanderify~\cite{fommtool} is a wrapper around first-order-model~\cite{fomm} that exposes a simple user interface for anyone with any level of technical skill.

The above works focus majorly on one problem, which makes them unsuitable for editing videos for translation. We present a video editing tool with multiple SOTA deep learning features available with one click of a button, allowing easy and high-quality video editing.

\section{Talking Face Video Editing} \label{Talking face features}

Visual presentation of a talking person requires generating image frames showing the speaker in various views while pronouncing various phonemes. The creation of realistic content synthetically has been a very active field. While most of these works have been published and often the implementations are also publicly available, these works are far from being widely used in real-world applications. The generation of a talking face consists of many sub-problems that are attempted separately without connecting the sub-problems. We observe that the automated versions are unsuitable for video editing due the editor's lack of control. For starters, the standalone algorithms work on entire videos. However, in real-world situations, these algorithms need to be used often in short temporal segments of a larger video. The algorithms may also work on spatial crops of a video instead of the whole frames. In such cases, the larger video needs pre-processing, which involves cropping and trimming the portion of interest and processing it. The editor then needs to insert the processed output back into the original video. To avoid using external pre-processing options, which are tedious and hard to integrate, we provide the option to the user to select the segment which needs to be processed interactively. We also provide extensive manual controls for trimming, cropping, and splitting videos. Similar options are also provided for the audio to select portions of audio to be fed into the required algorithms. We provide multiple state-of-the-art features in our editor to generate talking face features and generate realistic results by using them with our tool. Please note that these features can be accessed interactively via the GUI interface. The results generated are provided in the supplementary videos.

\begin{figure*}[h]
  \includegraphics[width=\linewidth]{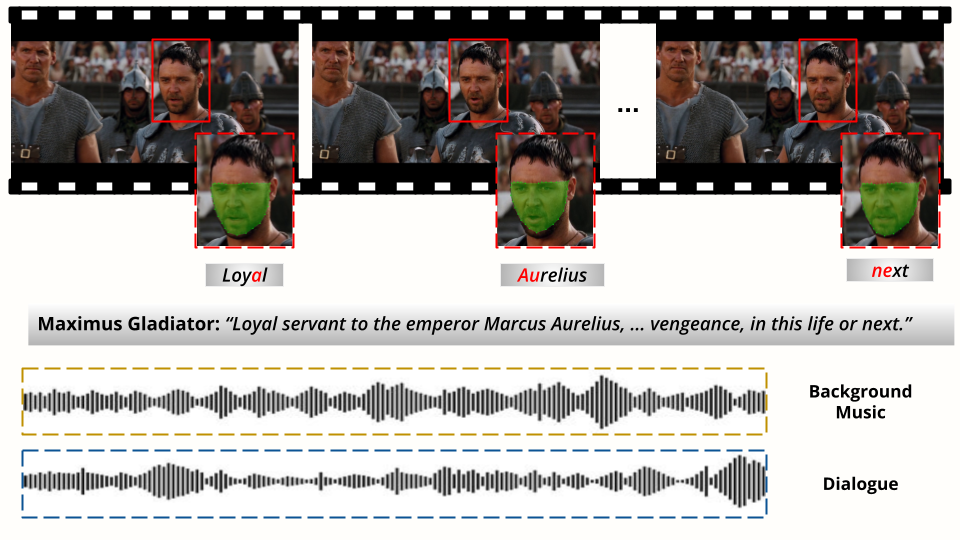}  \caption{Demonstration of the lipsync module working with the tool. The lipsync module processes the region in the bounding box drawn by the user. For better results, we only replace the facial region and not the entire bounding box enclosed by the facial key points to reduce the artifacts. The background music is separated from the audio to improve the result further.}
  \label{fig:gladiator}
\end{figure*}

\subsection{Generating Talking Heads from Audio} \label{MakeItTalk}
Generation of expressive talking-head videos from a single facial image with audio as the only guiding input is a challenging problem that can transform film-making and video editing in many ways. Algorithms like \cite{zhou2018talking,obama,nspvd,song2019talking} have tried to tackle this problem.
The main challenges include synchronizing audio and facial movements and generating multiple talking heads conveying different personalities. Finally, such algorithms need to generate lip-synced talking faces along with the complete set of head motion, lip motion, expressions only with audio, and an identity image as input.

We choose the current state-of-the-art~\cite{makeittalk} because of its ability to disentangle the speech content and speaker identity information in the input audio signal. It is then used to generate lip movements, facial movements corresponding to the expression, and the rest of the head poses. This disentanglement leads to plausible results and is considered one of the best systems around for this problem. Furthermore, the system was publicly released and is based on the PyTorch~\cite{NEURIPS2019_9015} library making it easier to integrate. 

We add this feature to our tool for various reasons. In translation systems, there is often a mismatch of length between the audio and video. When audio length exceeds the video's length, \citet{makeittalk} can be used to fill the required gap by generating a synthetic video of the speaker. Furthermore, the same can be extended to education lectures where the professor is not visible in the slide while teaching. This improves the user experience by a huge margin. Using MakeItTalk for video editing can lower the camera recording time and potentially reduce studio costs.

The user can select the audio segment in our tool by placing the markers over start and end positions. The user can place the cursor at the frame to select the images used for this feature. After the source image and audio are selected, the user needs to click on the \textit{MakeItTalk} button, and the video generated gets displayed in the tool. The generated video is also stored temporarily. The user can place the generated video at the required timestamp and then export the final video.


\subsection{Lip-syncing Talking Face Videos to a given Audio} \label{Wav2Lip}
 
Lip-syncing talking faces for given audio requires modifying lip movements without changing other characteristics like head pose and emotion.
Algorithms like \cite{yousaidthat, lipgan,wav2lip} are designed to morph the lip movements of speakers according to a guiding speech.
These algorithms input a speech segment and a talking face video segment of any identity to morph the speaker's lip movements, matching the input speech with high accuracy.

We choose Wav2Lip~\cite{wav2lip} due to its higher quality and superior performance. It is significantly more accurate than the previous works on this problem that can handle any arbitrary speech and identity. Furthermore, Wav2Lip works for videos in the wild, making it highly usable in various situations. One of the prominent use-cases of Wav2Lip is in lip-syncing dubbed movies and translated lectures. However, the automatic code-base again proves inadequate for editing complex movie scenes, and manual control becomes necessary.

\paragraph{Which portions of the video to lip-sync?} Lip-sync is required only at the places where a face is present in the video. To select that portion of the video, the user can mark the starting and ending timestamp of the video by placing the markers. The user can lip synchronize the segment by clicking the \textit{Wav2Lip} button. Like the MakeItTalk feature, the generated output is stored temporarily and can replace the non-lip synchronized segment.

\paragraph{Which face to lip-sync?} The Wav2Lip~\cite{wav2lip} lacks a mechanism to pick out a single face from multiple faces that can be lip-synced. To tackle this problem, the user can click the \textit{Select a face} button. The user can select a bounding box around the required face in the video frame, generating better outputs and decreasing inference time.

\paragraph{Pasting the resulting face crops} The existing algorithms replace rectangular regions around the face, which works decently in most cases.
Still, the artifacts are significantly visible when used for movies, and the output quality is reduced.
To solve this, we detect the face region using facial keypoint detection and replace only that portion of the face in the video as illustrated in \autoref{fig:wav2lip_fig}, producing lesser artifacts and higher quality output.

\begin{figure*}[h]
  \includegraphics[width=\linewidth,height=200px]{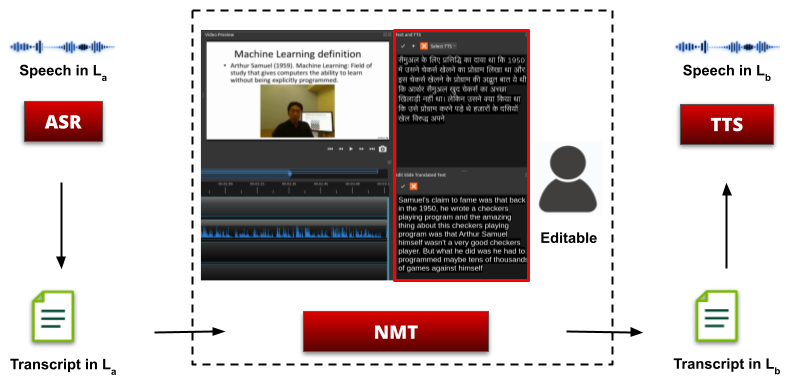}  \caption{Block diagram of speech to speech translation. The transcript in language L$_A$ is obtained from the speech using the ASR module, and then the text generated is translated into language L$_B$ using the NMT module, which is editable by the user. The speech in language L$_B$ is obtained from the translated transcript using the TTS module.}
  \label{fig:S2S}
  \vspace{0.3cm}
\end{figure*}

\subsection{Face Re-enactment to a Guiding Video} \label{Image Animation}
The task of face re-enactment consists of synthesizing videos automatically by extracting appearance from a source image and motion patterns from a driving video. Works like~\cite{gans,balakrishnansynthesizing,bansal2018recyclegan} have tried to tackle this problem; however, they all depend on pre-trained models to extract the information like keypoint locations. 

Monkey-Net by \citet{Siarohin_2019_CVPR} was the first object-agnostic model for image animation but suffered from poorly modeled object appearance transformation. This issue was tackled in~\cite{fomm}, which proposes using a set of self-learned key points and local affine transformations to model complex motions and claims to outperform state-of-the-art image animation methods significantly.

We choose to include~\cite{fomm} in our tool for numerous reasons. The generation of visual content by animating objects in still images serves many applications in movie production.
The translation of educational lectures serves as an essential application of content-synchronization (discussed in Section~\ref{Content Sync}).
In cases where translated audio is longer than its corresponding video, this feature can be used to generate video frames to sync the audio and video.
The synchronization is required for a short segment of video where the mismatch is present. Still, the standalone algorithm takes an entire video, and hence the processing of a short segment becomes a tedious task.
In our tool, we tackle this by allowing the user to select a driving video segment which provides the motion.
The user selects this video segment by placing two markers, one at the start and the other at the end. The user also places the cursor at the frame, which is selected as the source image. After the selection is made, the user can click on the \textit{FOMM} button to generate the result. After the result is generated, it is displayed in the tool, and the user can easily drag this segment to fill the gap between the audio and video segments.

\subsection{Editing a dubbed movie using Wav2Lip}

OTT content these days is the most significant source for entertainment, with every significant streaming platform having a plethora of options in many languages. However, a user is not free to consume whatever content they prefer because of the language barrier. Two heavily used methods to address this issue are to either view subbed content or dubbed content. The subbed content relies heavily on the user's ability to continue reading the dialogues while also focusing on the video, which is not ideal. In the dubbed content, the unsynchronized lip movements make the content unnatural. To address these issues, we show how our tool can be used to generate translated content with lip sync.

\autoref{fig:gladiator} shows a movie clip where multiple faces are present with a single speaker speaking in presence of background music. Wav2lip~\cite{wav2lip} is unable to produce quality results in such cases mainly because of two reasons:
  \begin{itemize}
      \item In the presence of multiple faces, there is no mechanism to select a face to be lip synchronized.
      \item The presence of background music makes it harder to generate accurate lip movements.
  \end{itemize}

The user can select the speaker by drawing a bounding box around the facial region to overcome the first problem. Wav2lip~\cite{wav2lip} then generates the lip movements for the speaker selected by the user. This selection reduces the search area for face detection resulting in faster and accurate inferences.
For the second problem, the user has an option to separate dialogues from background music~\cite{vocali}. Using only the separated dialogue as speech input for Wav2lip~\cite{wav2lip} generates more accurate lip movements.

\section{Translating Lectures from Language L$_A$ to Language L$_B$ with a Human in the Loop} \label{SST}

Given a lecture video in language L$_A$, we aim to translate the lecture in language L$_B$. Our tool provides an interface consisting of different modules using which a user can easily translate the lecture from one language to another. We begin with the translation of speech which follows the steps discussed by~\citet{lipgan}. However, we improve the automatic system by including manual control and editing capabilities, as shown in \autoref{fig:S2S}, in a systematic manner that leads to better quality translation. The length of the translated speech often varies from the video length, leading to out-of-sync content, corrected using the tool as discussed in Section~\ref{Content Sync}. The additional challenge of background content translation for lectures having slides is discussed in Section~\ref{Bg content sync}.

\begin{figure*}[h]
  \includegraphics[width=\textwidth]{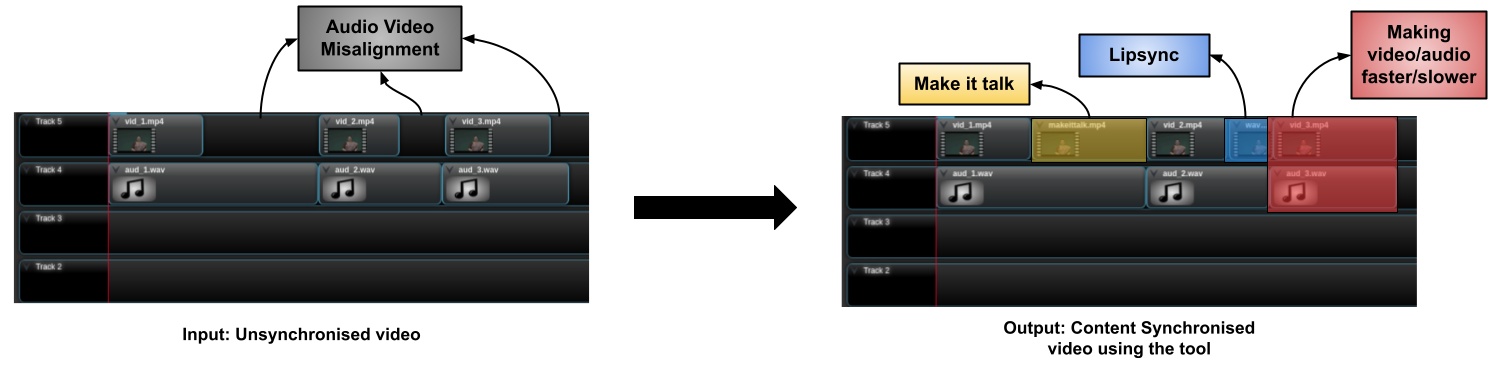}  \caption{illustrates the different features like generation of talking heads from audio, lipsync and speed manipulation of audio/ video present in the tool used to synchronize a misaligned video.}
  \label{fig:content_sync}
  \vspace{0.3cm}
\end{figure*}
\subsection{Automatic Speech Recognition} \label{ASR}
Automatic speech recognition (ASR) is the capability that enables machines to process human speech into a written format. Speech recognition is an essential module in video translations. There has been significant research in the area of automatic speech recognition. We use popular works on Automatic Speech Recognition (ASR)\cite{google,kranz_1986,amodei2015deep,collobert2016wav2letter,SileroModels} in our tool to obtain the corresponding text from speech.

We directly use transcripts when they are available (often present in YouTube) for a lecture and provide the user with an option in the tool to use Google's ASR~\cite{google} or Amazon Transcribe~\cite{kranz_1986} when the transcripts are unavailable. While these services are highly accurate and dependable, they are accessible only for a trial period; hence we also have integrated an open-source ASR~\cite{SileroModels} into the tool. ~\citet{SileroModels} can handle four languages: English, German, Spanish and Ukrainian. \citet{SileroModels} works reasonably well on any 'in the wild' audio with sufficient SNR and reports WER of 11.5 on LibriSpeech test set. We also provide users the option to edit the text manually. This helps the user correct transcription errors caused by the ASRs and correct even spoken language mistakes by the original speaker. We generate the output from the ASR of the user's choice and display it in an editable text box. The editor can modify the textual annotations, which is then fed to the NMT module discussed in Section~\ref{subsec:NMT}. 

\subsection{Neural Machine Translation} \label{subsec:NMT}

Neural Machine Translation(NMT) is an end-to-end learning approach that involves translating text from one language to another. The use of neural networks for NMT was first introduced in~\cite{sutskever2014sequence} using an end-to-end approach to sequence learning that makes minimal assumptions on the sequence structure. This was followed by the attention mechanism used in~\cite{bahdanau2016neural}, which improved the performance of such systems further. The rise of transformers~\cite{vaswani2017attention} has lead to massive gains since their introduction in 2017. This seminal work led to the meteoric rise of NMT, and soon hundreds of languages were covered. The Indian languages were also explored widely during this phase. Works like~\cite{jerin1, jerin2, jerin3} aimed to train models on major Indian languages and open-sourced them. Our editor includes Amazon Translate~\cite{amazon} and Google Translate~\cite{googlenmt} for translating English into other Latin languages. 
For translating Indian languages, we use the state-of-the-art system developed by~\citet{jerin3}. However, it is widely known that the best NMT systems are often unusable and require much manual correction. This is especially true when dealing with lectures that contain keywords that cannot be translated.   

We acknowledge this wide gap between the state-of-the-art NMT models and a human translator. We, therefore, provide the user with an option to edit the results from our NMT module. Similar to the ASR outputs, the NMT outputs are also displayed in an editable text box (Figure~\ref{fig:S2S}). The user can correct errors and ensure that the correct translations are passed to the text-to-speech module for generating the translated speech. The user can also optionally add delimiters like ``comma" and ``full-stop" to improve the naturalness of the generated speech. 
\begin{figure*}[h]
  \includegraphics[width=\textwidth, height=215px]{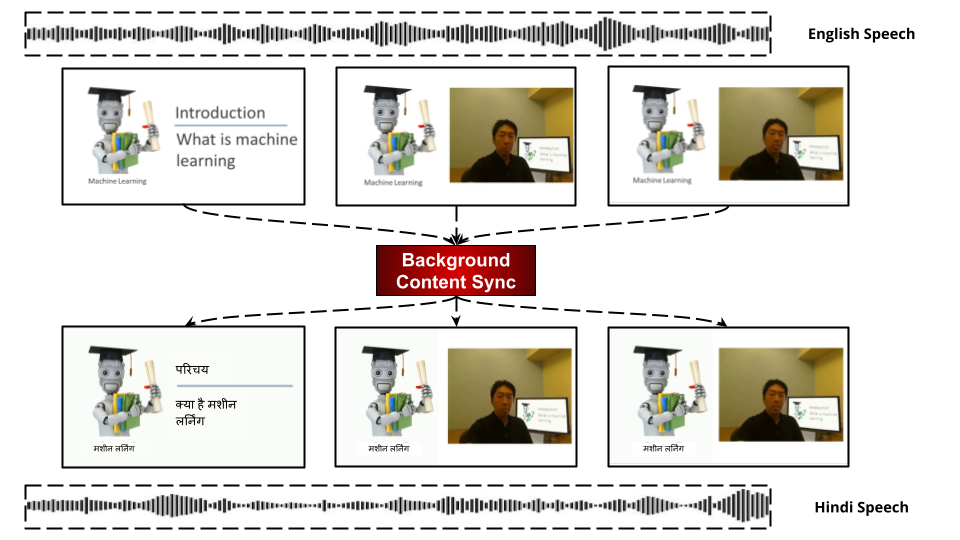}
  \caption{This figure shows the use of the Background Content Synchronization feature, using which the content of the slide in language L$_A$ is translated into language L$_B$. }
  \label{fig:bg_content_sync}
  \vspace{-0.3cm}
\end{figure*}

\vspace{-8pt}

\subsection{Text-to-speech} \label{TTS}

Text-to-speech (TTS) involves generating speech from text inputs. The field of text-to-speech has seen tremendous advancement in recent years. Several works \cite{tacotron2, deepvoice3, fastspeech, fastspeech2} have tried to tackle synthesizing speech with neural networks. Tacotron2~\cite{tacotron2}, a neural text-to-spectrogram conversion model uses Griffin-Lim for spectrogram-to-waveform synthesis. Deepvoice3~\cite{deepvoice3} uses a fully convolutional character-to-spectrogram architecture enabling parallel computation for faster training that competitive architectures using recurrent cells lack.
These works produce high-quality speech on the LJSpeech~\cite{ljspeech17} dataset and are comparable to the original human voice. But when trained on Indian regional languages, the results produced are far worse and deemed unusable. Another issue is the sub-par speech quality of the above-mentioned TTS while synthesizing results on longer sentences.

Thus in our tool, we use ~\citet{glowtts}, a flow-based generative model for parallel TTS that internally learns to align text and speech by leveraging the properties of flows and dynamic programming. Flow-based TTS' are robust and perform considerably better on longer sentences. We also find that it generalizes well on Indian languages compared to~\cite{deepvoice3,tacotron2}. The user can synthesize the speech directly from the translated text written in the text box by selecting the desired TTS model and clicking the \textit{TTS} button. We also provide a stand-alone option to synthesize speech by loading the text from a file.

\subsection{Content synchronization} \label{Content Sync}

While translating or dubbing a video, the speech and video often go out of sync, resulting in losing out on the experience of the original lecture. This causes a bigger problem in educational lectures as most educational lectures use slides to explain various concepts. Furthermore, if the speech and video are out of sync, the translated speech may end up getting overlaid to the wrong slides, causing significant issues in the viewer's understanding.

Editing a longer video is a tedious task; we segment the video into smaller chunks to make it simpler. Following this, each segment can be synchronized independently. In the end, the user can combine all the segments to generate the final result. The whole procedure is done interactively. The different features used for content synchronization of the audio/video segments have been illustrated in \autoref{fig:content_sync}. 


\begin{itemize}
    \item The user can manually place the audio or video at any timestamp to correct the sync.
    \item The user can alter the audio or video length by manipulating the speed of that segment to match the length of the corresponding audio or video segment.
    \item Using Section~\ref{MakeItTalk}, \ref{Wav2Lip}, and \ref{Image Animation} the user can generate synthetic video frames to fill the gap caused due to differences in audio and video length.
\end{itemize}

\subsection{Background Content Translation} \label{Bg content sync}

Background content translation involves translating the content present in the video in language L$_A$ to language L$_B$. It is an essential step in video translation. We use a combination of OCR and NMT modules for this task. 
OCR, or optical character recognition, is one of the earliest addressed computer vision tasks. Deep learning techniques \cite{tesseract-ocr,jaidedai,wojna2017attentionbased,baek2019wrong}  have significantly improved the accuracy and are used for industrial applications. Pytesseract~\cite{pytesseract} is an open-source python wrapper for Google’s Tesseract-OCR~\cite{tesseract-ocr} Engine used for this task.

We include LayoutParser~\cite{shen2021layoutparser}, a python library, in the tool, which is built on top of Pytesseract. LayoutParser provides an additional benefit of ``line detection" over Pytesseract's word-level detection, which helps improve the content translation.

Educational lectures heavily make use of slides for teaching. To explain a single slide, the instructor may take several seconds or even minutes, keeping the slide constant for the whole period. We exploit this prior in our tool for achieving background content translation. The user can select the video segment's starting and ending timestamp of constant slide/background in the editor. The user then gets the translations of the content into a text box by clicking on the \textit{Background Content Translation} button. The automatic translation is obtained from Section~\ref{subsec:NMT} and is improved by human translators via editing the translated text. After the user finalizes the text, the user can click the 'Overlay on slide' button to replace the translated text into the slide. This feature achieves results as shown in~\autoref{fig:bg_content_sync}.

\subsection{Combining the Steps to Translate a Lecture}

Translating a lecture by segmenting it into multiple smaller chunks is easier than entirely translating at once. Our tool significantly reduces the manual effort required to translate the lecture by enabling the user to process chunks individually.

The first step in translating a lecture from language L$_A$ to L$_B$ is to translate the speech from L$_A$ to L$_B$. The ASR module present in the tool helps get the speech's transcript in language L$_A$. The transcript is then translated into L$_B$ by the NMT module. With manual control, the user gets to correct the errors in automatic translations. The translated text is then converted into L$_B$ speech by using the TTS module. The output obtained from the TTS module in language L$_B$ is often different in length than the original speech in language L$_A$. This difference can result in misalignment of the audio and video segment of the lecture. To avoid such misalignment, the user can use MakeItTalk, FOMM, and speed manipulation features. With this, the user can achieve translated audio-video synchronized content. 

The user can also translate the lecture slides' content wherever present, from language L$_A$ to L$_B$ using Section~\ref{Bg content sync} with manual control to correct the translation errors due to automatic translation.
The final step in the translation is to lip synchronize the lecture with the output speech using the Lipsync feature.

\section{System Details}
OpenShot~\cite{openshotstudios} is a python based tool implemented in the PyQt library. We have added the above-discussed features in the tool, accessible to the user using a button click. For every feature, the audio and video segments selected by the user are extracted from the original video and are then processed. The tool takes the same amount of time to generate the outputs as the original works. The generated outputs are temporarily stored and displayed in the tool for that particular session, which can replace the original video segment or add to fill the gaps due to audio and video duration mismatch. The user can specify the quality and fps for saving the video. The time required to save the video is proportional to the duration and quality of the video, where an hour-long 720p video takes nearly 2 to 3 minutes to export. To make the process of installation easier for the user, we provide an alternate where the computation takes place on a remote server so that the user does not have to worry about the required deep learning packages.

\section{Evaluations}
In this section, we conduct two types of experiments to understand the impact of our tool on both the human editors and viewers who consume the final videos. We first understand how far our tool helps humans to edit videos in Section~\ref{subsec:manual}, followed by a comprehensive user study on $25$ participants to rate the quality of the generated outputs through manual editing. 

\subsection{Comparing Manual Effort}
\label{subsec:manual}
This section compares the time taken to edit the video with and without using our tool. Five video segments of $10$ minutes were edited by $5$ participants, (a) with and (b) without using our tool. We recorded the time taken in both cases for (i) Audio translation (ii) Background Content Translation (iii) Content Synchronization. The average time (in minutes) for these tasks is reported in \autoref{tab:comparison}. 

\begin{table}[h]
\footnotesize
\begin{center}
\begin{tabular}{ |c|c|c|c|c| } 

\hline
\textbf{Method} & \textbf{S2S} & \textbf{Video Editing} & \textbf{BG. Translation} & \textbf{Total}\\ 
\hline
External & $33.7 \pm 3.76$  & $48.1 \pm 2.81$ & $17.3 \pm 2.58$ & $99.1 \pm 5.61$ \\
\hline
\textbf{Our Editor} & \textbf{$6.8 \pm 0.94$} & \textbf{$11.9 \pm 1.05$} & \textbf{$6.2 \pm 0.3$} & \textbf{$24.9 \pm 1.68$}\\
\hline
\end{tabular}
\end{center}
\caption{ present the comparison of the time taken to translate lecture videos using our tool and stand-alone automatic systems + external editors. The time are mentioned in minutes. S2S: Speech-to-Speech translation, Video Editing: Using talking face features and general cropping/trimming actions, BG Translation: Translating the background content. The last column denotes the total time taken on average by the editors.}
\label{tab:comparison}
\end{table}

While using method (b), users spent maximum time on editing tasks like splitting videos into smaller segments, processing each segment using automatic algorithms, and loading the processed segments back into the tool.
Since these standalone algorithms provide no manual control, there is no option to correct the mistakes that occur at different stages of translating the lecture, resulting in inaccuracies. Method (a) provides an easy-to-use interface to correct the automatic translations, significantly reducing the time to get correct translations versus when not using our tool. The entire video translation process takes a huge amount of effort and time, which reduces considerably using our tool.

\subsection{User Study to Rate the Quality of Results}
\begin{figure}[h]
  \includegraphics[width=\linewidth]{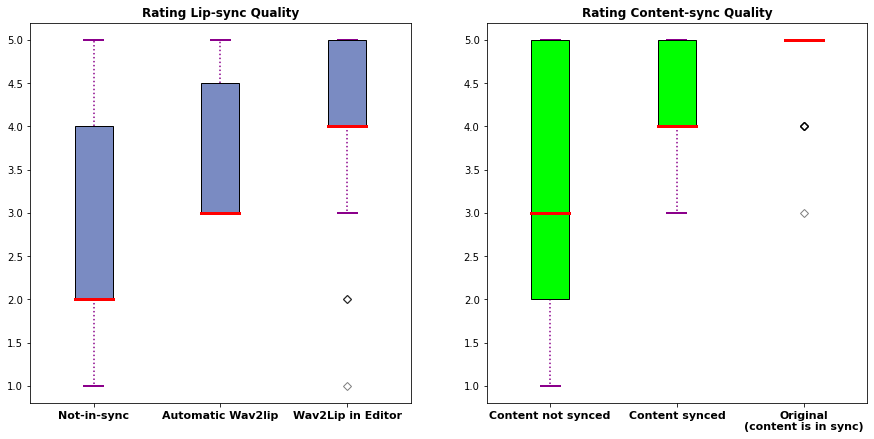}
  \caption{The boxplots show the distribution of values recorded from the experiments conducted. The left plot records the lipsync quality and the right plot records the content synchronization quality.}
  \label{fig:res}
  \vspace{-0.3cm}
\end{figure}

Users can use the tool to generate visual content for direct human consumption. Hence, we subject the results generated from the tool directly to human evaluation. The participants are asked to evaluate the quality of lipsync in lectures and dubbed movies. The participants were also asked to rate the translated lectures in terms of content sync. For both these tasks, we ask the users to rate between $1$ to $5$ where $1$ is the lowest rating, and $5$ is the best possible. $25$ users participated in our study. Male:Female ratio was $\approx 1$, and the participants were in the $21 - 30$ age range. 

We compare results from our editor with those from the automatic Wav2lip~\cite{wav2lip} algorithm and the out-of-sync original video. As observed from Figure~\ref{fig:res} (box plot on the left), the median user ratings for the output from our tool was $4$ out of $5$, while automatic Wav2Lip had a median rating of $3$. The not-in-sync output was rated the lowest.

For the second task, we use multiple videos to compare the original video, translated video by simply overlaying the translated speech, and translated video by using our tool. As observed from figure~\ref{fig:res} (box plot on the right), the median rating of our tool was $4$ out of $5$, while simply overlaying the translated text was rated at a median of $3$, and the original videos received a median score of $5$. 

Overall, we get conclusive evidence that our tool improves the time required for editing a video and reduces the manual effort while improving the results.

\section{Conclusion}
In this paper, we present a version of the OpenShot video editor with multiple new functionalities. We incorporate a speech-to-speech translation pipeline where automatically generated results can be corrected manually. We also include several state-of-the-art talking face generation techniques into our editor, editing complex dubbed movie scenes, and translating lectures into a more straightforward task. Our tool can translate background content on the slides and ensure the synchronization between the speech and the background content is preserved using systematic steps. We believe our editor opens up a new paradigm in video editing that allows interactive use of the latest published research and improves the overall user experience.

\bibliographystyle{ACM-Reference-Format}
\bibliography{acmart}

\appendix




\end{document}